\documentclass[conference]{IEEEtran}
\IEEEoverridecommandlockouts
\usepackage{cite}
\usepackage{amsmath,amssymb,amsfonts}
\usepackage{algorithm, algorithmic}
\usepackage{graphicx}
\usepackage{textcomp}
\usepackage{caption}
\usepackage{xcolor}
\def\BibTeX{{\rm B\kern-.05em{\sc i\kern-.025em b}\kern-.08em
    T\kern-.1667em\lower.7ex\hbox{E}\kern-.125emX}}

\newcommand{\E}{\mathrm{E}}
\newcommand{\Var}{\mathrm{Var}}

\begin{document}

\title{Counting and Algorithmic Generalization with Transformers}

\author{\IEEEauthorblockN{Simon Ouellette}
\IEEEauthorblockA{\textit{Corresponding author} \\
s.ouellette1@gmail.com}
\and
\IEEEauthorblockN{Rolf Pfister}
\IEEEauthorblockA{\textit{Lab42 AI Research Institute} \\
 Davos, Switzerland}
\IEEEauthorblockA{\textit{Munich Center for Mathematical Philosophy} \\
Ludwig-Maximilians-Universität München}

\and
\IEEEauthorblockN{Hansueli Jud}
\IEEEauthorblockA{\textit{Lab42 AI Research Institute} \\
 Davos, Switzerland}
}

\maketitle

\begin{abstract}
Algorithmic generalization in machine learning refers to the ability to learn the underlying algorithm that generates data in a way that generalizes out-of-distribution. This is generally considered a difficult task for most machine learning algorithms. Here, we analyze algorithmic generalization when counting is required, either implicitly or explicitly. We show that standard Transformers are based on architectural decisions that hinder out-of-distribution performance for such tasks. In particular, we discuss the consequences of using layer normalization and of normalizing the attention weights via softmax. With ablation of the problematic operations, we demonstrate that a modified transformer can exhibit a good algorithmic generalization performance on counting while using a very lightweight architecture.
\end{abstract}

\section{Introduction}

Algorithmic generalization and extrapolation are machine learning functionalities that are considered difficult to achieve. They are, however, essential capabilities of human intelligence. For example, the Abstraction \& Reasoning Corpus (ARC) challenge \cite{ARC} is a set of visual reasoning tasks intended to be a test of intelligence. It requires the ability to learn the algorithm that solves a task from very few examples. Doing so requires a capability for abstraction and for reasoning, as its name implies. In particular, notions of cardinality are one of the many requirements to solve the ARC challenge (example in Appendix A).

Although research in this field has produced some interesting breakthroughs, such as the Neural GPU \cite{neuralGPU1, neuralGPU2, neuralGPU3} and the Universal Transformer \cite{UT1, UT2}, there remains a lot of work to do. While both approaches have increased performance levels on algorithmic tasks relative to prior work, they still fall well below perfect generalization performance, and struggle on more complex tasks.
In this paper we use counting as an example to demonstrate the types of failure conditions that can occur when attempting to learn algorithms with Transformers. To be more specific, we point out architectural decisions underlying the standard Transformer that hinder cardinality-based generalization. 

In particular, we identify layer normalization and scaling of the attention weights via softmax as the two main operations that make it impossible for standard Transformers to learn counting in a generalizable way. Both architectural choices rescale values and thereby assume that quantity (absolute value) is irrelevant. Normalization is a common technique in machine learning, because it tends to smooth out gradients and it helps with stability and convergence, especially for deeper architectures. 
However, in doing so, they force the neural network to learn a mapping function from natural quantities to normalized numerical vectors that overfits the training set distribution. In contrast, if we let the natural quantities speak for themselves, we find that the model generalizes much better out-of-distribution.

The decisions behind the Transformer architecture are validated when we are manipulating fundamentally qualitative entities, as in natural language processing tasks. However, with the increasing popularity of Transformers, attempts are made to apply them to other types of tasks. The insights presented here should be carefully considered when doing so.

Our contributions are two-fold:
\begin{enumerate}
\item We show that standard Transformers are architecturally unable to learn counting in a generalizable manner.
\item We identify a previously unexplored problem in layer normalization with respect to out-of-distribution generalization.
\end{enumerate}

\section{Related work}

\subsection{Algorithmic generalization with Transformers}
Algorithmic tasks include problems such as learning multi-digit multiplication, sorting lists, and string manipulation. What is common to these types of tasks is the need for iterative, exact processing of intermediate steps according to some rules (which must be learned). Research on algorithmic tasks especially places emphasis on out-of-distribution generalization. The latter implies that the correct algorithm was learned, as opposed to a mere memorization of the training domain.

Standard Transformers \cite{transformer} are architecturally limited to a static number of sequential operations \cite{UT1}. By construction, only \emph{N} sequential attention+feed-forward operations can be applied to each token, where \emph{N} is the number of encoding and decoding layers. Thus, they lack the recurrent inductive bias, which appears crucial for robust generalization when the training and test set differ in required processing depth.

To address this limitation, Universal Transformers were developed \cite{UT1}. They are composed of a standard encoder and decoder block over which a potentially dynamic number of iterations can occur, determined by the Adaptive Computation Time mechanism. This concept relies on a separate neural network that determines the halting condition, thereby enabling conditional loops. The authors show that it outperforms the standard Transformer on a variety of algorithmic and reasoning tasks. The Universal Transformer has been further enhanced with the addition of an auxiliary grid-like memory module \cite{UT2}, thereby enabling new levels of algorithmic generalization on the multi-digit multiplication problem.

Other research \cite{SA_limitations} demonstrates mathematically that standard Transformers "cannot model periodic finite-state languages, nor hierarchical structure, unless the number of layers or heads increases with input length". Indeed, once again, the static number of layers or heads is shown to hinder learning of processes that require arbitrary amounts of iteration or recursion.

\subsection{Counting}

It has been shown that recurrent neural networks, especially Long Short-Term Memory (LSTM) networks, are capable of some level of counting generalization \cite{LSTM-counting, counting2}. However, it is also notable that there is always a certain degree of performance degradation whenever the test set goes beyond the scale of the training set \cite{counting3}.

Algorithmic tasks include so-called counter languages: formal languages that implicitly require some degree of counting ability. For example, a simple counter language is Dyck, in which the alphabet is composed of only opening and closing parentheses. Roughly described, it is a set of balanced strings of parentheses, where each opening parenthesis has a corresponding closing parenthesis in the correct order. Thus, "()(())" is a well-formed sentence in the Dyck language, while ")()((" is not. The machine learning task on these counter languages is usually to distinguish well-formed from illegal sentences.

Transformers can learn some counter languages (like Shuffle-Dyck and n-ary Boolean Expressions), but they perform worse than LSTMs overall. Out of 7 formal counting languages, LSTMs generalized perfectly on all of them, while Transformers failed to generalize on 3 of them \cite{NLP_numeracy2}.

Transformer-based NLP models are able to answer numerical reasoning questions from the DROP dataset, suggesting a certain degree of emergent numeracy. When inspecting the embeddings, it was found \cite{NLP_numeracy2} that pre-trained embeddings contain fine-grained information about cardinality and ordinality. However, a significant degradation of performance has been observed when the model needs to extrapolate to numbers beyond the training range.

\section{Methodology}

\subsection{Two counting paradigms}

We postulate that there are two fundamental ways in which a Transformer can learn to count items in its input sequence. There is a parallel method, and an iterative (or sequential) method. That standard Transformers cannot count iteratively is a direct consequence of prior work \cite{UT1, SA_limitations}: they can only perform a fixed maximum number of sequential operations. This is further detailed in the section "Counting the Iterative Way".

However, there is no \emph{a priori} reason to believe that the attention mechanism is unable to simultaneously attend to all identical objects and sum them up in one operation. To better understand the parallel method, we conceptualize Algorithm \ref{alg:algorithm}, which represents a plausible algorithm for parallel counting of pixels in a grid. From there, we show that it is possible to map a single Transformer encoder layer onto that algorithm, but only if certain modifications are made. In order to explain the algorithm, we must first explain details of the problem to be solved.

In our counting experiments, we have as input a sequence representing a grid (which is flattened). Each token in the input sequence corresponds to a cell pixel in the grid. Each pixel can have one of ten different colours, which is one-hot encoded over a 10-dimensional vector. As a result, for a 6x6 grid, for example, we have a corresponding matrix of dimension [6, 6, 10], which is in turn flattened to a sequence of [36, 10].

The goal of this experiment is to output a sequence that contains the count of instances of each color in the grid. The output is of the same dimension as the input and each cell contains the number of occurrences of that cell's color. More specifically, it is the dimension of the one-hot encoding that was set to 1 in the input grid that will contain this count in the output grid.

For example, suppose that in the 10-dimensional one-hot encodings the second dimension represents the color blue, and that the input grid contains 5 blue pixels. Then, each blue cell in the output grid will be encoded as: [0, 5, 0, 0, 0, 0, 0, 0, 0, 0]. Since the original intention is to count "non-background" colors, the color zero (which is the black color, i.e. the background color) has ground truths that are always set to 0, regardless of how many instances it has.

With this in mind, we can explain the parallel counting algorithm. The loop iterates over tokens in the input sequence (in a Transformer this can be parallelized). The flattened input grid $M^{I, J}$ has two dimensions: the number of pixels $I$ and the color encoding $J$. Each token $m_i$ is a pixel or, in other words, a color vector.

The vector $\textbf{w}$ (of length $I$) contains a value of $1$ where the token's color in $M^{I, J}$ corresponds to the current color $m_i$. Otherwise, the vector element is set to $0$. The operation $V = \textbf{w} \cdot M$ is the matrix-vector product between $\textbf{w}$ and $M$, i.e. the dot-product of each element $w_i$ with each row $m_i$.

This gives the matrix $V$ of the same shape as $M$. $count$ is the sum over the rows of $V$, with $count$ being a vector of the same length $J$ as the one-hot encoding. Thus, $count$ will contain the total number of instances of color $c$ in the grid $M$, in the $c^{th}$ dimension of $count$. All other dimensions will be 0. We then concatenate this to the output matrix, which will be of the same shape as $M$.

\subsection{Scaled dot-product attention}

Scaled dot-product attention (Appendix B) is introduced in \cite{transformer}. To better understand why this prevents counting, we performed the experiments \emph{Std-Transformer-Count} and \emph{No-LayerNorm-Count} \footnote{Code for all experiments in this paper can be found at: https://github.com/SimonOuellette35/CountingWithTransformers}. 

In \emph{Std-Transformer-Count}, we use a standard Transformer encoder module with only 1 layer and train it on randomly generated data for this task.
In \emph{No-LayerNorm-Count}, we disable its softmax operation over the attention weights, as well as its layer normalizations. The training procedure is the same as for \emph{Std-Transformer-Count}.

As detailed in the Experiments section, \emph{Std-Transformer-Count} fails to learn even in-distribution. \emph{No-LayerNorm-Count} reaches perfect performance in-distribution, and it also generalizes well to grid dimensions not seen during training. It does so by learning to count in the parallel way, essentially learning a mathematically equivalent form of Algorithm \ref{alg:algorithm}.

\subsection{Layer normalization}

Layer normalization (Appendix C) is motivated by the fact that it helps with training convergence by stabilizing gradients. However, in the name of smoothing out the gradients, we lose key information: absolute values. In other words, we lose information about quantities. This is in the same spirit as the softmax operation over the attention weights: the assumption is that only the relative numerical values across the various dimensions matter, not absolute quantities themselves. 

Thanks to the learned $\gamma$ and $\beta$ parameters in equation \ref{eq:layernorm} (Appendix C), it is still possible to output unnormalized values, such as count quantities, for example. That is, the normalized value of a dimension can be multiplied by an arbitrarily large coefficient $\gamma$ or added to an arbitrarily large bias $\beta$, resulting in any arbitrary value that does not need to be constrained between -1 and 1.

Experiments \emph{LayerNorm-SA-Count}, \emph{LayerNorm-FF-Count}, \emph{LayerNorm-Identity} and \emph{No-LayerNorm-Identity} are all intended to empirically support, as well as better analyze, this phenomenon.
In \emph{LayerNorm-SA-Count} and \emph{LayerNorm-FF-Count}, we use the same counting task as in the previous experiments. Starting from the modified Transformer model used in \emph{No-LayerNorm-Count}, we re-introduce the layer normalization operation in two steps. In \emph{LayerNorm-SA-Count}, we enable it at the level of the self-attention (SA) module, while keeping it disabled in the feed-forward (FF) network module. In \emph{LayerNorm-FF-Count}, we keep it disabled in the SA module, while re-enabling it in the FF network.

In \emph{LayerNorm-FF-Count}, the FF network is essentially trying to learn the identity function (because the SA module itself is enough to learn the counting task). We will show with experiments \emph{LayerNorm-Identity} and \emph{No-LayerNorm-Identity} that layer normalization hinders out-of-distribution generalization even for the identity function. In these experiments, rather than using as inputs and outputs integer values that represent counts, we use floating-point numerical vectors.

Specifically, for the training set, we generate 5-dimensional vectors whose numerical values are randomly picked in the range [-0.5, 0.5]. For the test set, however, we increase that range to [-1, 1]. The function to learn is the identity function: the ground truth and the inputs are the same values.

In support of the aforementioned theory that layer normalization essentially tethers the model to the statistical distribution of the training set, we show that a feed-forward network without layer normalization (\emph{No-LayerNorm-Identity}) can learn the identity function in a way that generalizes well out-of-distribution. However, once we add a layer normalization operation after its output (\emph{LayerNorm-Identity}), performance falls abruptly.

\begin{algorithm} [tb]
\caption{Generalizable parallel counting algorithm}
\label{alg:algorithm}
\textbf{Input}: $M$, the matrix representing the grid\\
\textbf{Output}: per cell pixel counts
\begin{algorithmic}[1] 
\STATE $count\_values \gets \{\}$
\FORALL {$m_i \in M^{I,J}$}
\STATE $w_i \gets 1\ if\ n_i=m_i,\ for\ n_i \in\ M^{I,J},\ 0\ otherwise.$
\STATE $V = \textbf{w} \cdot M$
\STATE $count_j = \sum_i{V_{i,j}}$
\STATE $count\_values \gets count\_values \oplus \textbf{count}$
\ENDFOR
\STATE \textbf{return} $count\_values$
\end{algorithmic}
\end{algorithm}

\subsection{Counting the iterative way}

The experiments show that a single-layer standard transformer encoder cannot learn counting in a parallel way. However, this does not rule out that a multi-layer transformer might learn it in an alternative way: via iteration. One such possibility is maintaining an internal counter, updated sequentially across its decoder layers. For example, suppose we structure the output sequence differently than in our previously discussed experiments: instead, each token in the output refers to a count for one for the 10 colors. Then, as the decoder iterates over the target sequence, the new position indicates what color to count. Based on this, each decoder layer looks up a pixel in the input grid, and increments the internal counter if that pixel's color is the one currently being counted.

However, even if such an algorithm works, prior work \cite{UT1, SA_limitations} makes it clear that a standard Transformer cannot learn an iterative algorithm that generalizes past a scale that is bounded by the fixed number of layers. If a Transformer has \emph{N} decoder or encoder layers, then it can perform at most \emph{N} sequential operations. While this would be enough to count up to approximately \emph{N} (or perhaps a fixed multiple thereof), there will always be a count \emph{N+1} that is unattainable by the architecture.

\section{Experiments}

Experiments \emph{Std-Transformer-Count}, \emph{No-LayerNorm-Count}, \emph{LayerNorm-SA-Count} and \emph{LayerNorm-FF-Count} are trained over 300k epochs at a learning rate of 0.0002 and with a batch size of 50. 

For experiments \emph{No-LayerNorm-Count}, \emph{LayerNorm-SA-Count} and \emph{LayerNorm-FF-Count} the FF network component is merely a single linear layer, because this was found to give better generalization results than the default 2-layer ReLU network.

In all experiments, we train the models on randomly generated grids of dimension 1x1 to 6x6 inclusively. We evaluate them on grids of 6x6 (in-distribution) and larger out-of-distribution grids. In models \emph{Std-Transformer-Count}, \emph{LayerNorm-Identity} and \emph{No-LayerNorm-Identity}, the FF network has 2 hidden layers of dimensionality 2048 each. The activation function used is a rectified linear unit. This is the default architecture of the FF module on the standard Transformer encoder.

The accuracy metric used to evaluate model performance rounds up the output matrix to the nearest integers, and checks if all of the rounded up values are exactly the same as the ground truth.

\paragraph{Std-Transformer-Count}
In Table \ref{tab:counting}, we see that the accuracy stays around 28\% regardless of the grid size. This suggests that the standard Transformer encoder architecture with 1 layer is intrinsically incapable of learning counting.

\begin{table*}[t]
\caption{Results for the experiments on the counting task}
\label{tab:counting}

\begin{center}
\begin{tabular}{
 | p{4.cm} | p{1.2cm} | p{1.2cm} | p{1.2cm} | p{1.2cm} | p{1.2cm} | p{1.2cm} | p{1.2cm} | p{1.2cm} | 
}
\hline
\textbf{Model} & \textbf{6x6} & \textbf{7x7} & \textbf{8x8} & \textbf{9x9} & \textbf{10x10} & \textbf{12x12} & \textbf{15x15} & \textbf{20x20} \\
\hline
Std-Transformer-Count & 32.48\% & 26.74\% & 25.10\% & 24.56\% & 25.39\% & 27.37\% & 31.12\% & 30.76\% \\
\hline
\textbf{No-LayerNorm-Count} & \textbf{100\%} & \textbf{100\%} & \textbf{100\%} & \textbf{100\%} & \textbf{100\%} & \textbf{100\%} & \textbf{100\%} & \textbf{100\%} \\
\hline
LayerNorm-SA-Count & 99.97\% & 99.80\% & 98.52\% & 92.59\% & 80.86\% & 45.99\% & 30.17\% & 33.79 \% \\
\hline
LayerNorm-FF-Count & 97.49\% & 90.07\% & 75.49\% & 56.05\% & 40.52\% & 27.98\% & 28.30\% & 34.11\% \\
\hline
\end{tabular}
\end{center}
\end{table*}

\paragraph{No-LayerNorm-Count}
In stark contrast to \emph{Std-Transformer-Count}, not only does this modified Transformer encoder layer learn to solve the task on the training set, it generalizes with perfect accuracy up to 20x20 grids (in fact, up to 100x100 grids, in our experiments).

\paragraph{LayerNorm-SA-Count}
Table \ref{tab:counting} indicates that this model has learned fairly well to solve the task on the training set. Yet, there is a relatively rapid drop in performance as we increase the size of the grids beyond that of the training set.

\paragraph{LayerNorm-FF-Count}
The performance is similar, but slightly worse, than \emph{LayerNorm-SA-Count}.

\begin{figure}[h]
\centering
\includegraphics[scale=0.26]{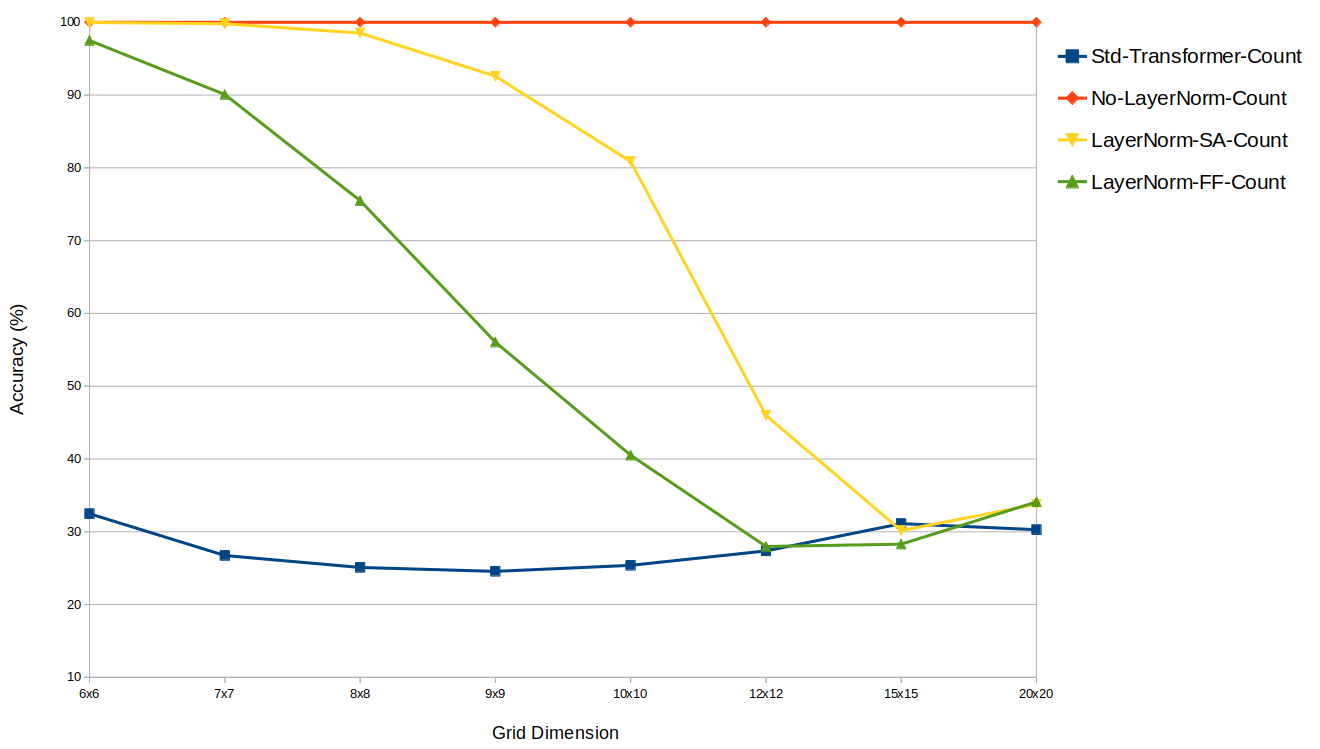}
\caption{Generalization performance (\%) comparison of different models}
\label{fig:counting}
\end{figure}

\paragraph{LayerNorm-Identity}
We train this model on the "identity task" that consists of 5-dimensional numerical vectors randomly generated in the interval [-0.5, 0.5]. Then, we test it on data of same dimension, but the interval is [-1, 1]. The goal is to simply learn the identity function: the predictions must reproduce the inputs. Because the outputs are continuous values, rather than integers or classes, we use the mean squared error as the loss function and evaluation metric. \emph{LayerNorm-Identity} was trained over 200k epochs, with a learning rate of 0.0001 and a batch size of 200. No further training was done because it was obvious from the learning curve that learning had stagnated.

\paragraph{No-LayerNorm-Identity}
In this variant of the experiment we disable layer normalization. The architecture, training procedure and task are otherwise identical to \emph{No-LayerNorm-Identity}, except that it was trained for only 100k epochs due to faster convergence. In Table \ref{tab:identity}, we see that the model without layer normalization significantly outperforms the model that uses layer normalization. 
 
\begin{table}[!htp]
\caption{Mean losses (std. dev. in parenthesis) for learning the identity function}
\label{tab:identity}
\begin{center}
\begin{tabular}{
 | p{3.5cm} | p{1.7cm} | p{1.7cm} | 
}
\hline
\textbf{Model} & \textbf{Train. loss} & \textbf{Test loss} \\
\hline
\textbf{No-LayerNorm-Identity} & \textbf{7.98e-08 (4.698e-08)} & \textbf{8.1954e-05 (1.279e-05)} \\
\hline
LayerNorm-Identity & 0.0174 (0.0019) & 0.1147 (0.0069) \\
\hline
\end{tabular}
\end{center}
\end{table}

\section{Discussion}

\subsection{Scaled dot-product attention}

There is a simple algorithm (Algorithm 1), that can solve the counting task in a way that generalizes beyond the training set distribution.

The first step in the loop consists of attending, for each cell in the grid, to all other cells in the grid that contain the same color. $\textbf{w}$ is the attention weights vector in which the weights are set to 1 whenever the color is the same as for cell $m_i$. These attention weights are multiplied by the input grid in order to get a vector where all cells with the same color have value 1 in the corresponding dimension, while the other cells have all zeros. By summing these up, we obtain the desired answer: the count of cells with the same color.

However, if we introduce a softmax on the weights $\textbf{w}$, instead of having weights of 1, we will have weights of $\frac{1}{d}$ where $d$ is the number of instances of the same color. Once the attention module sums up these cells to get the final attention output, we end up going back to the value 1. In other words, no matter how many cells of the same color there are in the grid, the result of the attention output is 1. 
So, this generalizable solution is not possible when a softmax operation is applied on the attention weights, as in the standard Transformer encoder. We theorize that this is why \emph{Std-Transformer-Count} fails to solve the counting task.

In \emph{No-LayerNorm-Count}, the model ends up inferring the attention schema displayed in Table \ref{tab:attn_schema1} when given the input sequence [6, 6, 0, 3, 7, 0, 4, 3, 6] representing a 3x3 grid.

\begin{table}[!htp]
\caption{Inferred attention weights in \emph{No-LayerNorm-Count}, for sequence [6, 6, 0, 3, 7, 0, 4, 3, 6]}
\label{tab:attn_schema1}
\begin{center}
\begin{tabular}{
 | p{.6cm} | p{.4cm} | p{.4cm} | p{.4cm} | p{.4cm} | p{.4cm} | p{.4cm} | p{.4cm} | p{.4cm} | p{.4cm} |
}
\hline
 \textbf{9-D seq.} & \textbf{[0]} & \textbf{[1]} & \textbf{[2]} & \textbf{[3]} & \textbf{[4]} & \textbf{[5]} & \textbf{[6]} & \textbf{[7]} & \textbf{[8]} \\
\hline
\textbf{[0]} & .23 & .23 & .00 & .00 & .00 & .00 & .00 & .00 & .24 \\
\hline
\textbf{[1]} & .23 & .23 & .00 & .00 & .00 & .00 & .00 & .00 & .24 \\
\hline
\textbf{[2]} & .00 & .00 & .00 & .00 & .00 & .00 & .00 & .00 & .00 \\
\hline
\textbf{[3]} & .00 & .00 & .00 & .22 & .00 & .00 & .00 & .22 & .00 \\
\hline
\textbf{[4]} & .00 & .00 & .00 & .00 & .23 & .00 & .00 & .00 & .00 \\
\hline
\textbf{[5]} & .00 & .00 & .00 & .00 & .00 & .00 & .00 & .00 & .00 \\
\hline
\textbf{[6]} & .00 & .00 & .00 & .00 & .00 & .00 & .23 & .00 & .00 \\
\hline
\textbf{[7]} & .00 & .00 & .00 & .22 & .00 & .00 & .00 & .22 & .00 \\
\hline
\textbf{[8]} & .23 & .23 & .00 & .00 & .00 & .00 & .00 & .00 & .24 \\
\hline
\end{tabular}
\end{center}
\end{table}

As can be seen, for each cell in the grid, a weight $\lambda = .23$ (approximately) is attributed to each other cell that contains the same color. At the same time, a weight of approximately 0 is attributed to the unrelated colors.
For example, cell 0 has color 6, which is found at indices 0, 1 and 8 of the sequence. Consequently, row 0 in the weight matrix has non-zero values at columns 0, 1, and 8. Note that the rows corresponding to the color zero do not contain any non-zero weights, simply because the model must learn to always output zero for the color zero.
Mathematically, then, the vector output $\hat{y}$ of each cell after the self-attention module is:

\begin{equation}
    \hat{y} = \sum_i{\lambda \cdot c_i} = \lambda \sum_i{c_i}
\end{equation}

\noindent where $c_i$ are the values of the cells that have the same color. One can immediately see that all we need to do to get the final answers, is to divide by $\lambda$. This is trivial for the FF network to learn. This learned model is mathematically equivalent to Algorithm 1, and generalizes well regardless of the count values or of the number of tokens.

In \emph{Std-Transformer-Count}, however, the Transformer encoder block fails to learn this model, because of the softmax operation that occurs in the standard Transformer's attention module. It effectively turns the previous formula into the following:

\begin{equation}
    \sum_i{\frac{1}{N} \cdot c_i}=\frac{1}{N} \sum_i{c_i} = \frac{\hat{y}}{N}
\end{equation}

\noindent where N is the number of same-color cells $c_i$. This is because, with the softmax operation, the attention weights must sum up to 1.
The result, $\frac{\hat{y}}{N}$ could be salvaged if the FF network had the ability to divide by $\frac{1}{N}$. However, the FF network does not have access to the value N, which is dynamic on a grid-by-grid basis, since it processes tokens one at a time, rather than the grid as a whole. It, therefore, cannot generate the desired final answer.

\subsection{Layer normalization}

\begin{figure}[h]
\centering
\includegraphics[scale=0.18]{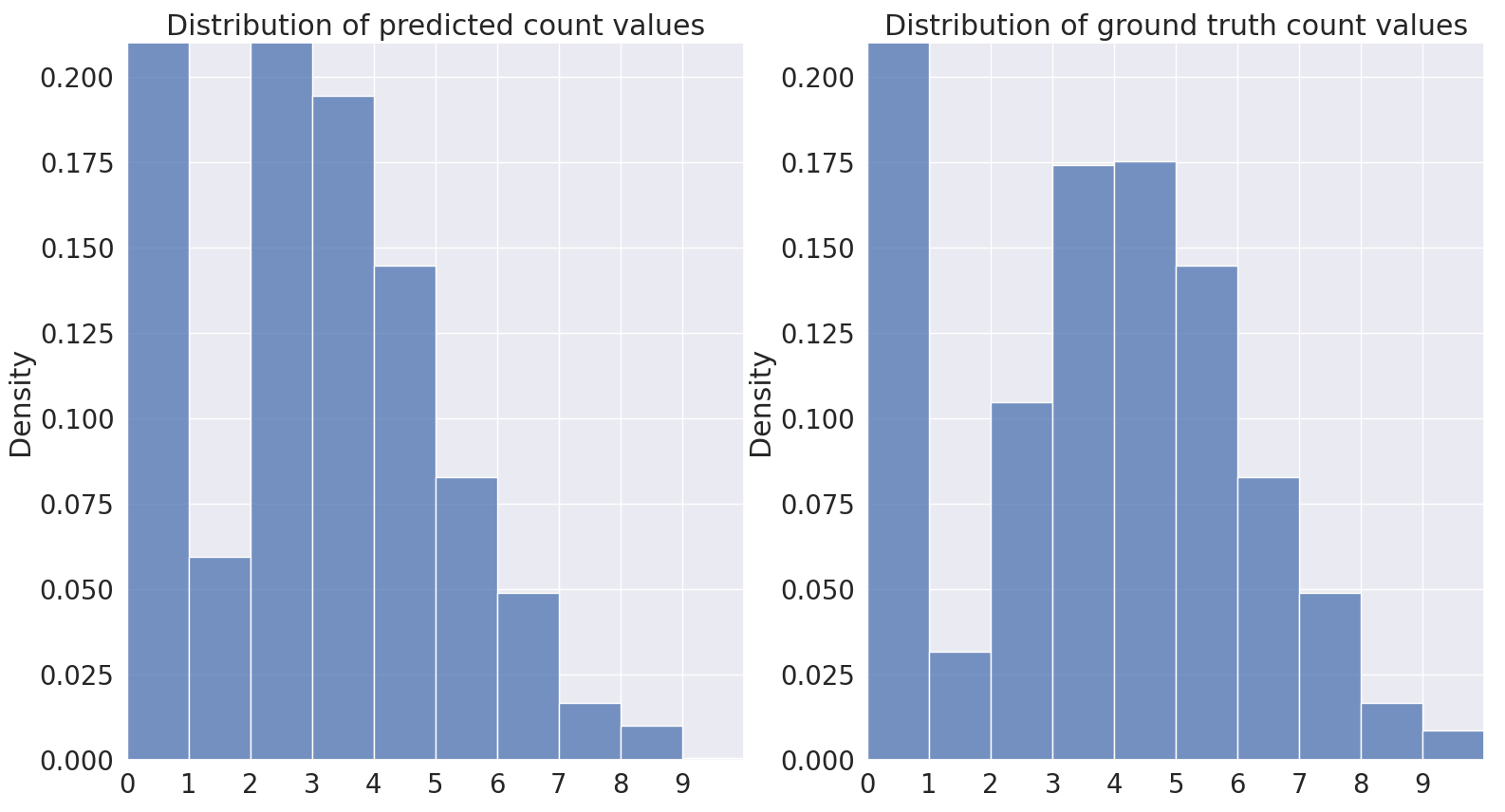}
\caption{Histograms of count predictions (left) vs ground truths (right) for 6x6 grids (LayerNorm-FF-Count)}
\label{fig:FF_count_hist1}
\end{figure}

\begin{figure}[h]
\centering
\includegraphics[scale=0.18]{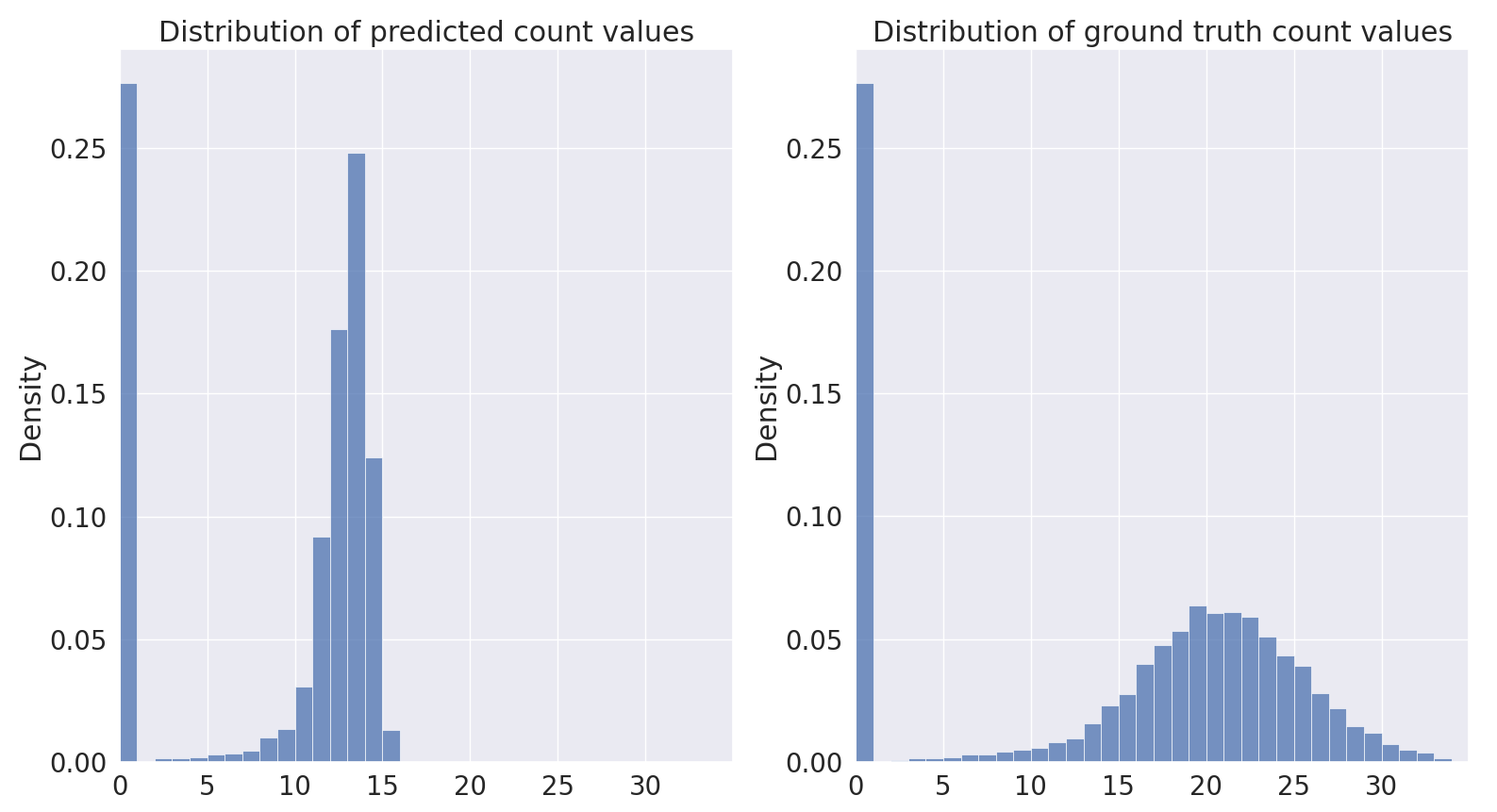}
\caption{Histograms of count predictions (left) vs ground truths (right) for 15x15 grids (LayerNorm-FF-Count)}
\label{fig:FF_count_hist2}
\end{figure}

The poor generalization performance of \emph{LayerNorm-FF-Count} can be directly observed in the differences between the distribution of the predictions and the ground truths. On the 6x6 grids, the prediction and ground truth count value distributions are approximately the same (see Fig. \ref{fig:FF_count_hist1}). However, on the 15x15 grids, where the error rate is high, the distributions are quite different (see Fig. \ref{fig:FF_count_hist2}). Note the high rate of zero-counts, due to the special case related to the background color. This artifact can be ignored for the purpose of this analysis.

We theorize that the FF neural network must anticipate the layer normalization operation and structure the inputs that it feeds to it such that the desired output survives the transformation. In particular, since the output of the layer normalization, in this case, is directly the count value prediction, the FF network must learn to "counter" the layer normalization operation. Successfully countering the layer normalization operation involves controlling the variance of the outputted vectors, because layer normalization divides by the standard deviation of a vector. It then multiplies each dimension of the vector by a learned coefficient (which is static once learned; it does not dynamically adapt to the vector itself). By controlling the output vector's variances, the model can ensure that the output of the layer normalization operation corresponds to the desired count value. 

\begin{figure}[h]
\centering
\includegraphics[scale=0.18]{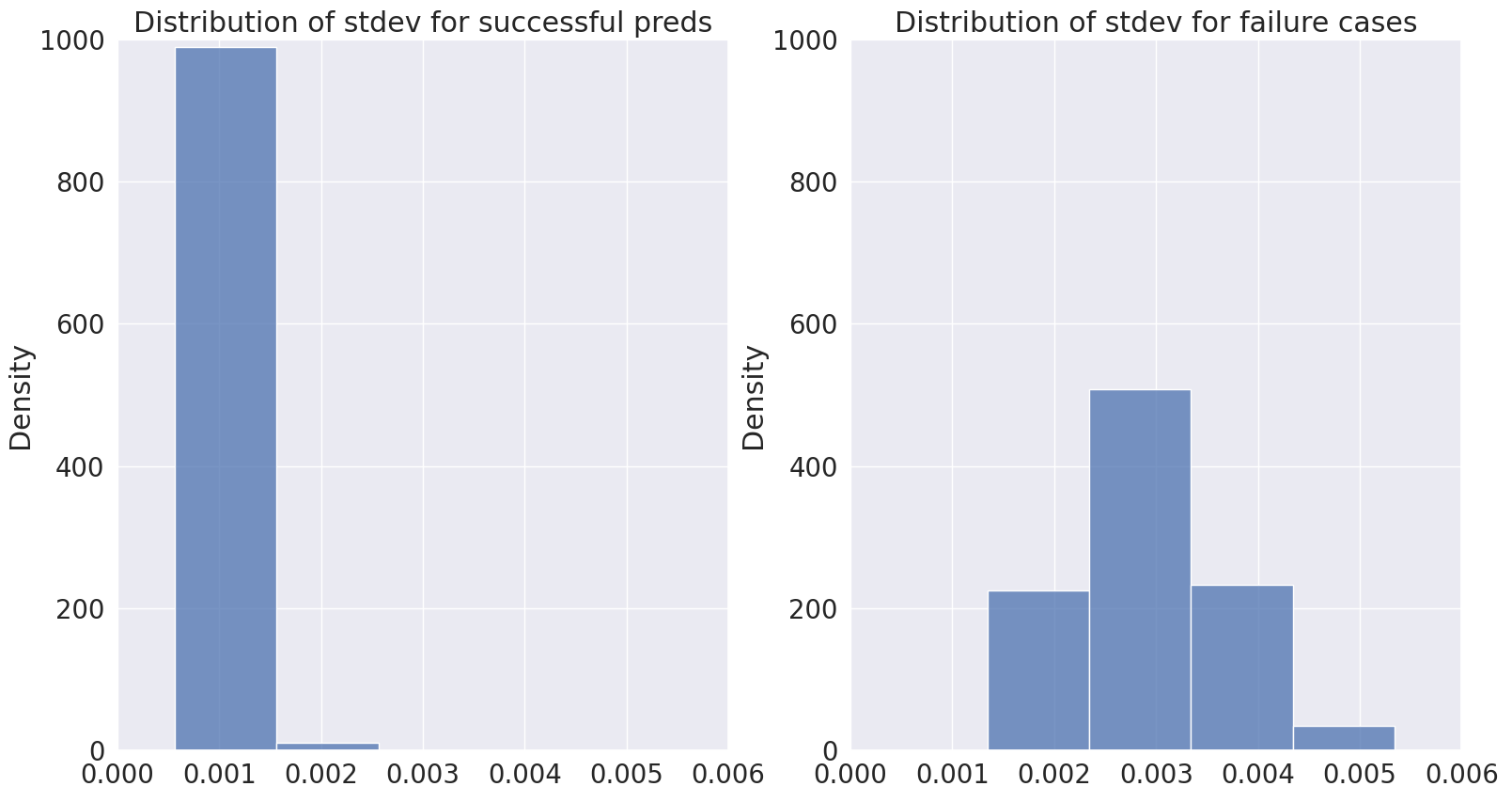}
\caption{Histograms of std. deviations for successful predictions (left) vs failed predictions (right) count values on 12x12 grids (LayerNorm-FF-Count)}
\label{fig:FF_stdevs}
\end{figure}

In Figure \ref{fig:FF_stdevs}, we see the standard deviations of the vectors outputted by the FF network (for 12x12 grids), before being passed to the layer normalization operation. When the standard deviation is close to 0.001, the prediction tends to be correct. When the standard deviation is higher than 0.001, it tends to be incorrect.

Mathematically, suppose that $\hat{y}$ is the intended count value prediction. Let $\gamma$ be the coefficient learned by the layer normalization for the corresponding dimension in the vector where $\hat{y}$ is located (remember that these count values are in fact 10-dimensional). Then, from the equation of layer normalization (see Appendix C, equation \ref{eq:layernorm}), the FF network must output a vector $\bar{v}$ containing value $\bar{y}$ such that:

\begin{equation}
    \frac{\bar{y} - \mathbb{E}[\bar{v}]}{\sqrt{\Var[\bar{v}] + \epsilon}} \cdot \gamma + \beta = \hat{y}
\end{equation}

\noindent A simple strategy, which our model learned, is to ensure that $\Var[\bar{v}]$ is approximately static (on the training set), at a value that counteracts the multiplication by $\gamma$. This way, the model can ensure that the output of the layer normalization operation corresponds to the desired count value.

The problem with this learned strategy is that the mapping from the FF network's input vectors (count values) to output vectors of a fixed variance evidently overfits the training set distribution. This is demonstrated by the significantly varying variances in the out-of-distribution data. These out-of-distribution variances, in turn, result in incorrect predictions once they pass through the layer normalization operation.

We see a similar phenomenon in \emph{LayerNorm-SA-Count}, where the self-attention linear output layer is forced to encode count values into normalizable numerical vectors. In other words, the information about the pixel counts must "survive" the normalization operation. This would not be possible by simply letting the count values speak for themselves, since the quantity is lost when subtracting by the vector mean and dividing by its standard deviation.

\begin{figure}[h]
\centering
\includegraphics[scale=0.18]{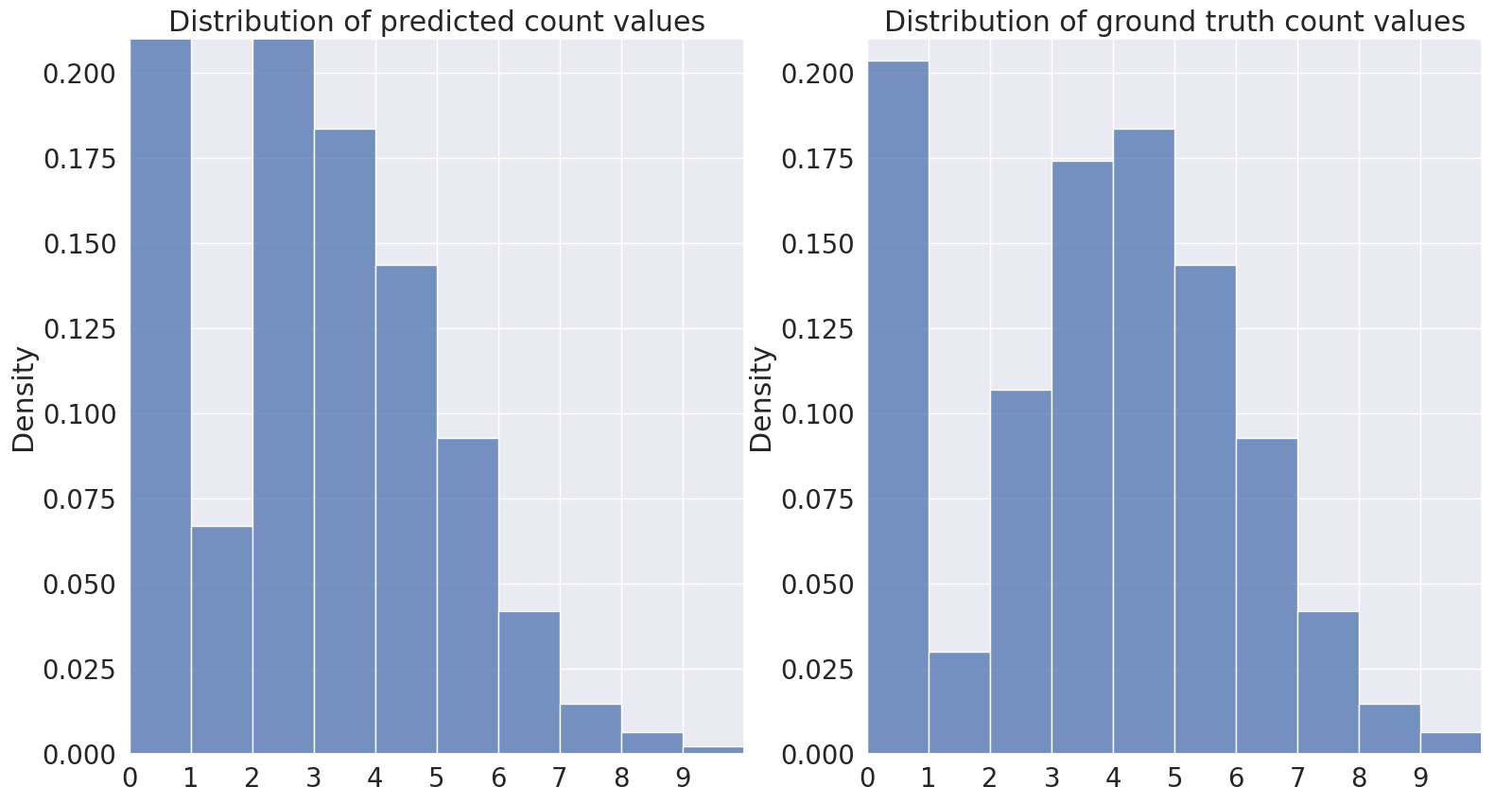}
\caption{Histograms of count predictions (left) vs ground truths (right) for 6x6 grids (LayerNorm-SA-Count)}
\label{fig:SA_count_hist1}
\end{figure}

\begin{figure}[h]
\centering
\includegraphics[scale=0.18]{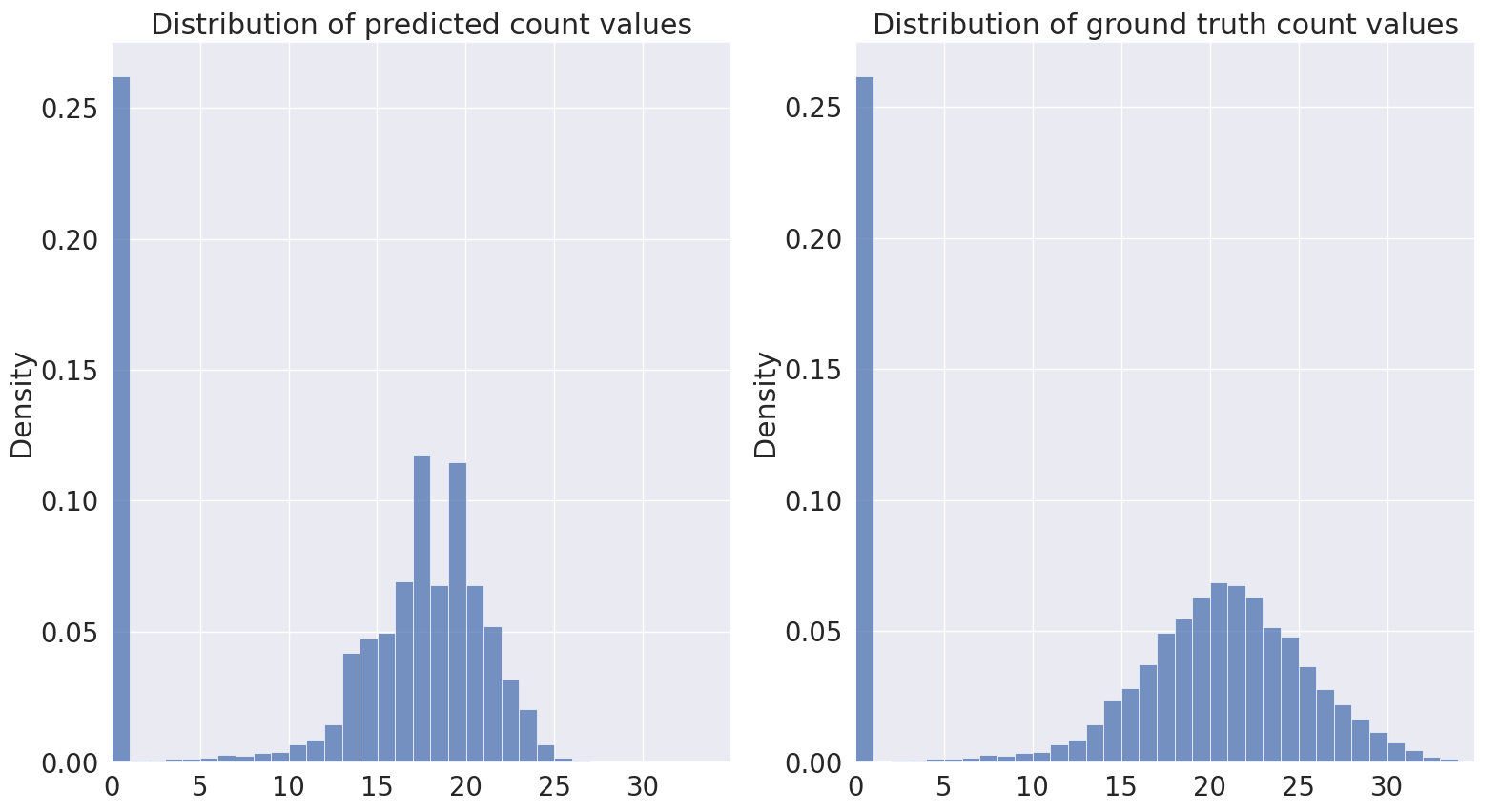}
\caption{Histograms of count predictions (left) vs ground truths (right) for 15x15 grids (LayerNorm-SA-Count)}
\label{fig:SA_count_hist2}
\end{figure}

\begin{figure}[h]
\centering
\includegraphics[scale=0.18]{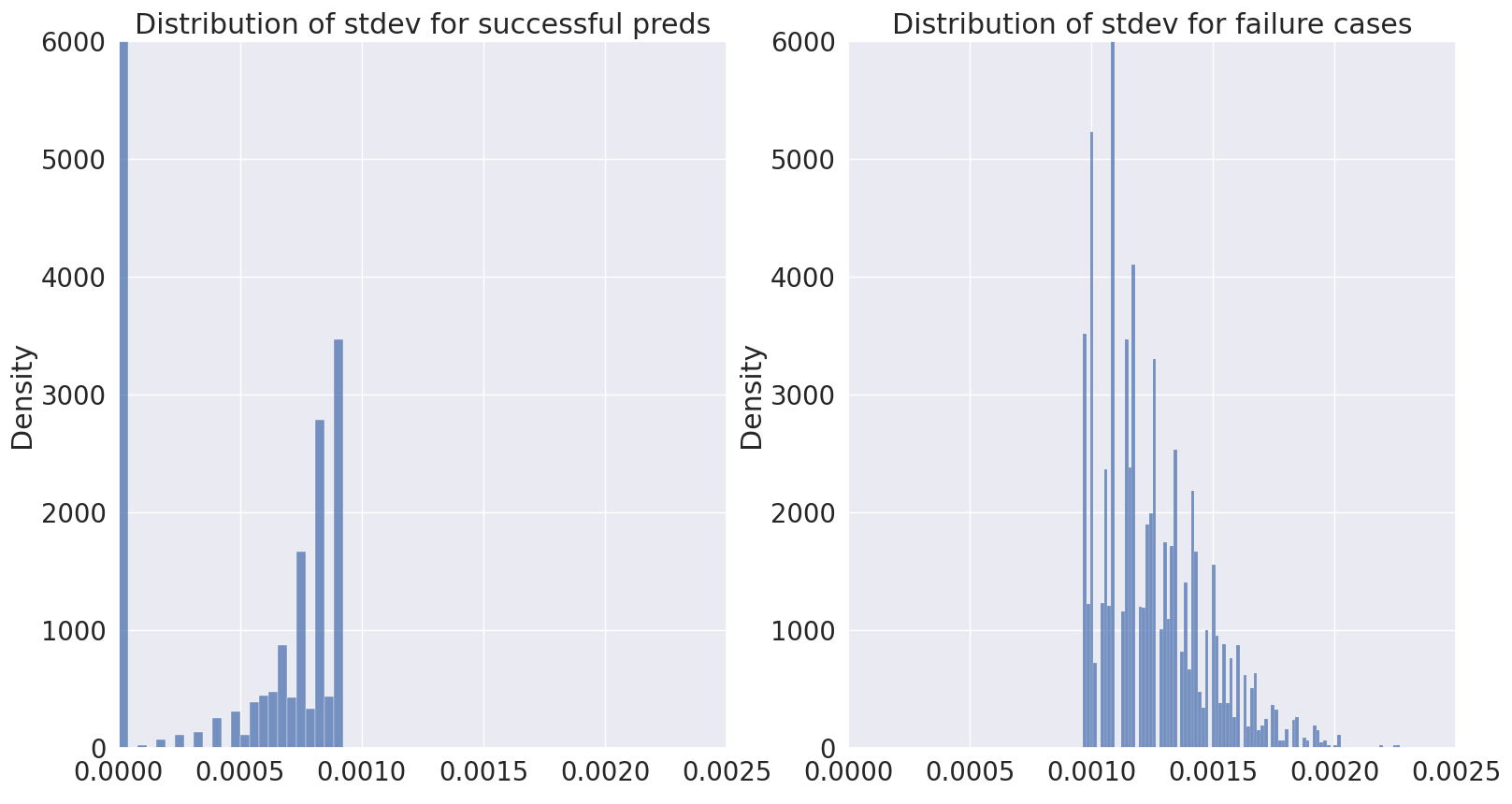}
\caption{Histograms of std. deviations for successful predictions (left) vs failed predictions (right) count values on 12x12 grids (LayerNorm-SA-Count)}
\label{fig:SA_stdevs}
\end{figure}

The impact on the distribution of predicted count values is more subtle in \emph{LayerNorm-SA-Count} than in \emph{LayerNorm-FF-Count} (see Fig. \ref{fig:SA_count_hist1} and \ref{fig:SA_count_hist2}). Here also, observing the standard deviations of the inputs of the layer normalization module is more informative (see Fig. \ref{fig:SA_stdevs}). We can see that inputs to the SA module that have a standard deviation below 0.001 are generally correct, while inputs to the SA module that have a standard deviation above 0.001 are incorrect.

\section{Conclusion}
Transformers are increasingly being used in research on algorithmic generalization. However, they were originally designed with a specific purpose that makes them well-suited to tasks like NLP. This is revealed in some architectural decisions that make fundamental assumptions about the un-importance of quantity. We show the consequences of these assumptions for algorithmic tasks, focusing on tasks that require explicit or implicit counting, and on the identity function. In particular, we point out that applying a softmax operation over the attention weights makes it impossible to learn to count entities across the input tokens in a parallel manner. Furthermore, we demonstrate that layer normalization causes models to overfit the statistical distribution of the training set. Further research is needed to determine the extent of this phenomenon, and to better understand it.

\bibliographystyle{ieeetr}
\bibliography{paper}

\section{Appendix}

\subsection{ARC task \emph{5582e5ca.json}}

\begin{figure}[h]
\centering
\captionsetup{justification=centering,margin=2cm}
\includegraphics[scale=0.4]{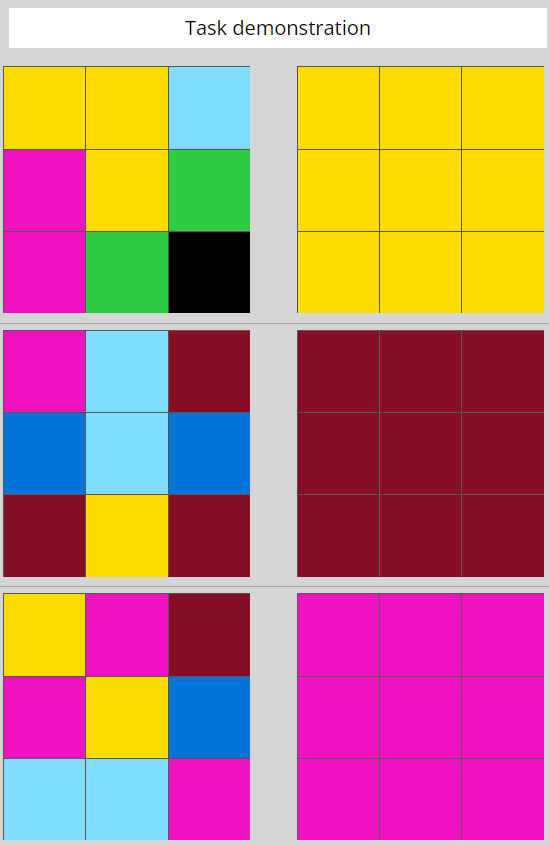}
\caption{ARC task 5582e5ca.json}
\label{fig:arc_task}
\end{figure}

In the ARC challenge, training task \emph{5582e5ca.json} consists of identifying the color that has the most instances in the input grid and to output it as a 3x3 grid (see Fig. \ref{fig:arc_task}). Learning to solve this task in a generalizable way implies being able to count the number of instances of each pixel color, and then selecting the maximum value out of those count results.

In Figure \ref{fig:arc_task}, we can see 3 examples of the task to learn. Each row is an example, where the left grid is the input, and the right right is the expected output. In this case, the grids have dimension 3x3, but different ARC tasks have different grid dimensionalities.

\subsection{Scaled Dot-Product Attention}

Mathematically, it corresponds to:

\begin{equation}
    \mathrm{Attention}(Q, K, V) = \mathrm{softmax}(\frac{QK^T}{\sqrt{d_k}})V
\end{equation}

\noindent in which $Q$, $K$, and $V$ are the Query, Key and Value matrices respectively. 

These inputs correspond to the relevant embeddings multiplied by their respective weight matrices $W^Q$ (the query weights), $W^K$ (the key weights) and $W^V$ (the value weights). These weight matrices are learned during training. The "relevant embeddings" fed to each of the 3 inputs (Query, Key, Value) depend on the context. For self-attention, the Query, Key and Value embeddings are the same. However, in the decoder block's second attention phase, Query and Key are the encoder's output, while Value is the output of the previous attention phase (which is self-attention to the target sentence).

\begin{figure}[h]
\centering
\includegraphics[scale=0.26]{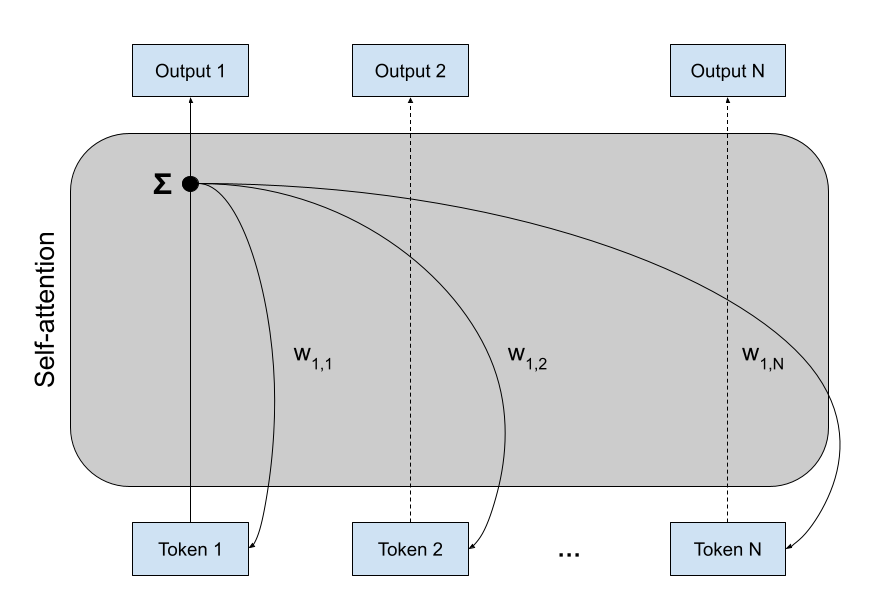}
\caption{Visualization of multiplication followed by summation in self-attention}
\label{fig:attention}
\end{figure}

They key element of interest here (for the main insights presented in the paper) is the softmax operation, which ensures that the attention weights sum up to 1 for each token in the input sequence. These attention weights $w_{i,j}$ are then multiplied by the token embeddings of the input sequence and summed up to obtain the output of the attention mechanism (see Fig. \ref{fig:attention}).

\subsection{Layer Normalization}

Layer normalization is mathematically described as:

\begin{equation} \label{eq:layernorm}
    y = \frac{x-\mathbb{E} [x]}{\sqrt{\Var [x]+\epsilon}} \cdot \gamma + \beta
\end{equation}

\noindent In contrast to batch normalization, layer normalization looks at each example from the batch separately, and normalizes (re-scales and re-centers) across its dimensions. $\E [x]$ is the average of values across a vector, while $\Var [x]$ is the variance across that same vector.

In addition to this operation, there are learned scaling parameters $\gamma$ and learned bias parameters $\beta$ for each of those dimensions, which are applied to the normalized vector.

\end{document}